\documentclass[conference]{IEEEtran}
\IEEEoverridecommandlockouts
\usepackage{cite}
\usepackage{graphicx,epstopdf,url}
\usepackage{subfigure}
\usepackage{verbatim}
\usepackage{textcomp}
\usepackage{amsmath}
\usepackage{amssymb}
\usepackage{multirow}

\usepackage{xcolor}
\usepackage{soul}
\usepackage{tabularx}
\usepackage{makecell}
\usepackage{wrapfig}
\usepackage{footnote}

\usepackage{booktabs}
\usepackage[utf8]{inputenc}  
\usepackage{indentfirst}
\usepackage[all]{xy}
\usepackage{bbm}
\usepackage{algorithm}
\usepackage{algpseudocode}
\usepackage{booktabs}

\usepackage{graphicx}

\newcommand{\our}{SPADE}

\def\BibTeX{{\rm B\kern-.05em{\sc i\kern-.025em b}\kern-.08em
    T\kern-.1667em\lower.7ex\hbox{E}\kern-.125emX}}
\begin{document}

\title{SPADE: Speculative Decoding for Precise and Low-Cost Distributed Edge–Cloud Inference}

\author{\IEEEauthorblockN{Divya Jyoti Bajpai, Kishan Kumar Upadhyay and Manjesh Kumar Hanawal}
\IEEEauthorblockA{\textit{MLiONS, Department of IEOR, IIT Bombay} \\
Mumbai, Maharashtra-400076, India\\
\{divyajyoti.bajpai, 24n0453, mhanawal\}@iitb.ac.in}
 }

\maketitle 

\begin{abstract}
Large Language Models (LLMs) have achieved remarkable success in natural language understanding and generation, but their deployment is constrained by high computational demands. Deploying smaller LLMs directly on the edge can circumvent this, but with degraded accuracy. Deploying smaller cloud-based big LLMs preserves performance, but at the cost of expensive per-token computation. We present a distributed inference framework, \our{}, that integrates speculative decoding (SD) across edge and cloud. A compact draft model deployed on the edge generates candidate tokens rapidly, and a large verifier model on the cloud validates these tokens in parallel. Accepted tokens are retained, while only rejections trigger verifier correction, substantially reducing the number of cloud queries. Our plug-and-play design shifts the bulk of computation to the edge, significantly lowers inference time and cloud cost, and preserves the accuracy of the big model without any retraining requirement. Our approach demonstrates a practical path toward scalable, cost-efficient, and accurate deployment of LLMs in real-world environments. Experimental results across multiple Natural Language Processing tasks using SpecBench and CNN/Dailymail datasets demonstrate that \our{} reduces the cloud model calls by $76\%$ with zero loss in accuracy as compared to the full model. \footnote{Anonymized code: \url{https://anonymous.4open.science/r/spade-anon-DC1F}}
\end{abstract}

\begin{IEEEkeywords}
Speculative Decoding, Distributed inference.
\end{IEEEkeywords}

\section{Introduction}

Large Language Models (LLMs) have achieved remarkable breakthroughs in understanding and generating language. Yet, this surge in model scale comes with a hidden cost: larger models demand vast computational resources and memory, which are rarely available on mobile or edge devices. As a result, deploying state-of-the-art models outside of high-performance cloud servers becomes a major bottleneck.

Existing literature has explored multiple strategies to alleviate this burden. Techniques such as model pruning \cite{michel2019sixteen}, weight quantization \cite{kim2021bert}, and knowledge distillation \cite{jiao2019tinybert} reduce the size and computational load of models, making them feasible for smaller devices. However, they often come at the cost of reduced accuracy, as simplifying the network limits its ability to capture complex patterns. Even purpose-built, smaller variants of large models, while more efficient, consistently underperform compared to their full-scale counterparts.

One alternative is to run the full model on the cloud. Hosting large models on powerful cloud servers can restore top-tier performance, enabling the use of models that are too large to run on mobile or edge devices. However, cloud-based inference introduces a key challenge: the computational cost grows with the duration for which the resources are utilized. As LLMs generate output in an autoregressive manner, they require sustained computation to process each input, which can result in longer inference times and increased costs. 

\begin{figure*}
    \centering
    \includegraphics[scale=0.7]{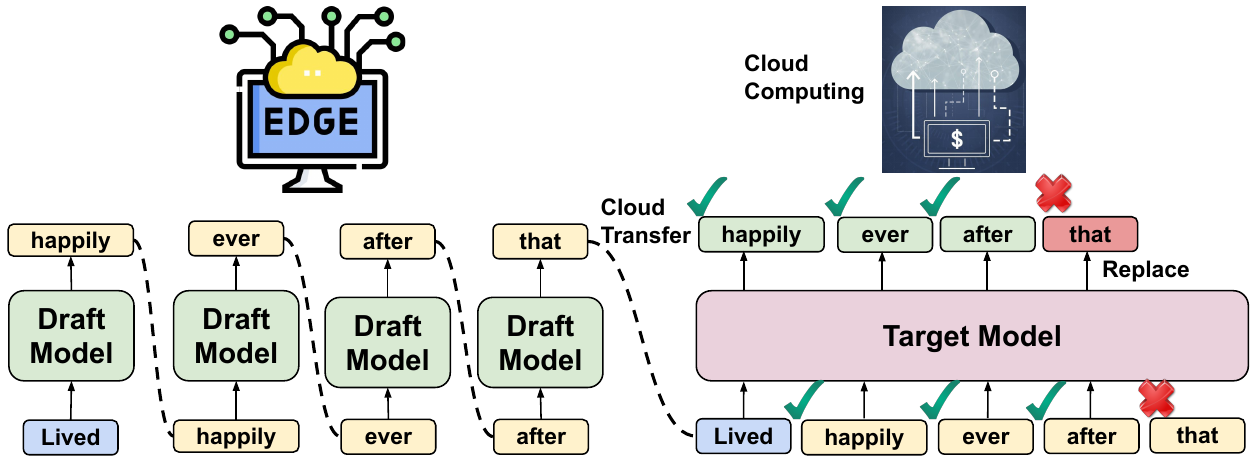}
    \caption{A lightweight edge model generates tokens autoregressively, which are sent to a cloud model for parallel verification and consistency checking. Accepted tokens are kept, rejected ones are replaced, and generation continues from the updated context.}
    \label{fig:main}
\end{figure*}

These constraints highlight a critical challenge: how can we achieve high accuracy without incurring large inference times on the cloud? This motivates the need for distributed inference between edge and cloud devices, where a smaller version of the LLM is deployed on the edge while the full-scale model is hosted in the cloud. Inference fully on the edge can reduce the accuracy, and inference fully on the cloud will increase the cost to a greater extent. The central question that remains is how to further reduce the reliance on the cloud by utilizing the resources available at the edge device, while still maintaining the accuracy of the full model.

Recently, Speculative Decoding (SD) has emerged as a powerful technique for accelerating LLM inference while preserving the accuracy of the full model. The approach relies on a two-model setup: a smaller, faster draft model that proposes speculative tokens and a larger, more accurate verifier model that evaluates them in parallel. Verified or accepted tokens are retained, while the first rejected token is replaced with the verifier’s predictions, after which the draft model continues generation conditioned on the corrected context. This iterative process cuts the autoregressive structure of LLMs to significantly reduce the expensive verifier calls—for instance, a $50$-token sequence that would normally require $50$ full model call can often be completed with only a few model calls—thus providing substantial efficiency gains without sacrificing output quality.

We propose \our{}, a distributed inference framework across edge and cloud devices, addressing the dual challenges of efficiency and accuracy in large-scale LLM deployment. In this framework, the smaller model deployed on the edge acts as a draft generator, producing a sequence of speculative (draft) tokens locally. These draft sequences are then transmitted to the larger, full-scale model hosted on the cloud, which serves as a verifier to evaluate and correct the discrepancies in a single pass. By generating the draft tokens at the edge and only verifying them in the cloud, our approach significantly reduces the computational burden on the cloud, decreases inference latency, and minimizes the number of expensive model calls on the cloud, while maintaining the output fidelity of the full-scale model without any training requirements. Our method is detailed in Fig.~\ref{fig:main} (details in Sec.~\ref{sec:SD-Distributed}).

The importance and usefulness of this method lie in its ability to enable scalable and resource-efficient deployment of large LLMs in real-world scenarios. It allows the edge device to handle most of the generative computation, reducing dependency on the cloud and improving the cloud computation cost. At the same time, the cloud ensures accuracy of the full model, so there is no compromise in output quality with minimal resource utilization of the cloud. This combination makes the deployment of large LLMs feasible in settings where purely cloud-based or purely edge-based solutions would be inefficient or impractical. In summary, our key contributions are as follows:

\begin{itemize}

    \item \textbf{Distributed Speculative Decoding Framework:} We introduce a novel edge-cloud distributed inference setup that leverages speculative decoding to efficiently split computation between a smaller edge model and a full-scale cloud model.
    
    \item \textbf{Reduce Cloud Computation:} Our method significantly reduces the number of calls to the cloud model by using as the edge model to generate the fraft sequences. This lowers the inference time and resource usage without compromising accuracy.
    
    \item \textbf{Maintain Full-Scale Accuracy:} The framework ensures that the final outputs are equivalent to those of the full cloud model, guaranteeing fidelity while benefiting from edge-side computation. This is achieved without any additional training requirements.
    
    \item \textbf{Empirical evaluations:} We provide empirical results on various NLP tasks using SpecBench and CNN/DailyMail datasets, where the cloud computation time was reduced by $76\%$ with no loss in performance.
\end{itemize}

\section{Related works}

We review prior works on distributed inference and speculative decoding related to our method.  

\noindent
\textbf{Distributed Inference:}  
Recent works leverage heterogeneous resources across edge and cloud devices to distribute computation.  
\emph{1) Layer-splitting:} Neurosurgeon \cite{kang2017neurosurgeon} partitions DNNs by executing initial layers on the edge and remaining layers on the cloud.  
\emph{2) Encoder-side training:} Methods improve edge encoders for efficient transmission, e.g., head-network distillation \cite{matsubara2021neural}.  
\emph{3) Early-exit classifiers:} Intermediate classifiers enable fast inference by exiting early for easy samples \cite{bajpai-hanawal-2024-ceebert}. SplitEE \cite{10.1145/3639856.3639873} and I-SplitEE \cite{10622954} optimize splitting and prediction.  
\emph{4) Complexity-aware routing:} DIMEE \cite{11161749} routes samples based on estimated complexity, but relies on dataset-specific heuristics.  

\noindent
\textbf{Speculative Decoding:}  
Speculative Decoding (SD) \cite{xia2024unlocking} accelerates autoregressive models using a lightweight draft model and a larger verifier for parallel validation.  
\emph{Self-Speculative Decoding:} LayerSkip \cite{elhoushi2024layer} reuse early layers of the large model as the draft.  

Distributed inference methods face key trade-offs:  
(i) lenient routing reduces performance,  
(ii) strict routing increases cost, and  
(iii) Complexity estimation lacks generalization.  

\textbf{Our Approach:}  
Our method differs as follows: 
\begin{itemize}
    \item We apply speculative decoding to distributed inference, with a draft model on the edge and verifier on the cloud.  
    \item We achieve \emph{zero performance loss} relative to the large model as proven in \cite{leviathan2023fast}.  
    \item Our framework generalizes across tasks and avoids dataset-specific heuristics.  
    \item SPADE is fully plug-and-play and requires no retraining.  
\end{itemize}

\section{Problem Setup}
We begin by introducing the general framework of autoregressive decoding, speculative decoding (SD), followed by our adaptation of SD for distributed inference.

\subsection{Autoregressive Decoding in LLMs}
LLMs typically adopt the transformer architecture, consisting of an embedding layer, $L$ stacked transformer blocks, and a final language modeling head. Let the input prompt be denoted by $y_{1:t}= (y_1, y_2, \ldots, y_t).$ 
The embedding layer maps input tokens to embeddings $x_0$. At each layer $l \in \{1,\ldots,L\}$, the hidden representation evolves as
$x_{l+1}^t = x_l^t + f_l(x_l^t),$ where $f_l(\cdot)$ denotes the transformation at layer $l$ (self-attention and feedforward). The final representation at the $L$-th layer, $x_L^t$, is projected to logits via the language modeling head, $o_L^t = g(x_L^t)$. Autoregressive decoding estimates the conditional distribution of the next token as
\[
P(y_{t+1} \mid y_{1:t}) = \mathrm{softmax}(o_L^t).
\]
The next token $y_{t+1}$ is then selected either via \emph{greedy decoding} (argmax) or \emph{sampling} for diversity.  

A key limitation of this procedure is its computational cost: generating each new token (word) requires a full forward pass of the model over the entire context. Consequently, inference latency is dominated by model depth and memory requirements, leading to inefficiencies in large-scale models.

\subsection{Speculative Decoding (SD)}
\label{sec:SD}
To mitigate the inefficiency of standard autoregressive decoding, \emph{SD} employs a two-model pipeline that combines a smaller, faster model (\emph{draft model}) with the full-scale model (\emph{verifier}). The procedure consists of two stages:  

\begin{enumerate}
    \item \textbf{Drafting Stage:}  
    The draft model autoregressively generates a block of $d$ candidate tokens,
    \[
    y_{t+1:t+d} = (y_{t+1}, y_{t+2}, \ldots, y_{t+d}),
    \]
    conditioned on the prefix $y_{1:t}$.  

    \item \textbf{Verification Stage:}  
    In a single forward pass, the verifier evaluates the block $y_{t+1:t+d}$ against its probability distribution. Tokens consistent with the verifier are accepted directly. The first token $y_j$ that is rejected is replaced by sampling from an adjusted distribution (explained next), and the draft model resumes generation from the updated prefix $(y_1, \ldots, y_j)$ where $j$ is the index of the token where the first rejection happens.
\end{enumerate}

This procedure ensures that the generated sequence is statistically identical \cite{leviathan2023fast} to what would have been produced by the verifier alone, while substantially reducing the number of expensive verifier calls. By verifying tokens in blocks rather than individually, SD achieves significant speedups in practice with minimal loss of performance. 

\subsection{Speculative Decoding for Distributed Inference}
\label{sec:SD-Distributed}
We utilize speculative decoding in a distributed inference framework that leverages both edge and cloud resources.  

\textbf{Draft Model $\mathcal{M}_q$ on Edge:}  
A compact LLM $\mathcal{M}_q$ is deployed on the edge device to perform fast inference under constrained memory and bandwidth. Model selection at the edge is governed by available resources such as GPU memory, CPU throughput, and system bandwidth. The draft model must be sufficiently lightweight to generate speculative sequences without overwhelming the device.  

\textbf{Verifier Model $\mathcal{M}_p$ on Cloud:}  
The cloud hosts a larger, high-accuracy model $\mathcal{M}_p$ that validates the candidate tokens proposed by $\mathcal{M}_q$. The computational demands of $\mathcal{M}_p$ make it unsuitable for edge deployment but ideal for cloud platforms, which can scale elastically with demand. Since cloud services often charge per model call, minimizing verifier invocations directly reduces inference cost.  

\textbf{Verification Criterion:}  
Let $p(x)$ denote the target distribution from $\mathcal{M}_p$, and $q(x)$ the draft distribution from $\mathcal{M}_q$. A candidate $x \sim q(x)$ is accepted with probability
$$\alpha(x) = \min\left(1, \frac{p(x)}{q(x)}\right).$$
If rejected, a replacement token is sampled from the adjusted distribution
$$p'(x) = \text{norm}\big(\max(0,\, p(x) - q(x))\big),$$
which corrects the bias introduced by the draft distribution. This guarantees equivalence to verifier-only decoding \cite{leviathan2023fast}. The approach is depicted pictorially in Fig~\ref{fig:main}.

\textbf{Pipeline:}  
A pseudo-code of our method is given in Algorithm \ref{alg:sd-distributed}. The procedure is summarized as follows: starting from a user prompt, the edge model $\mathcal{M}_q$ generates a block of $d$ draft tokens. These tokens are sent to the cloud, where $\mathcal{M}_p$ verifies them in parallel using a single forward pass. All tokens accepted under $\alpha(x)$ are appended to the output. The first rejected token is replaced via sampling from $p'(x)$, and the remaining draft tokens are discarded. The context is then updated with the accepted and corrected tokens, and the draft model resumes autoregressive generation for the next block until the end-of-sentence $<eos>$ is generated.  

This design guarantees that at least one token is appended to the sequence in each iteration, thereby ensuring progress. By batching verification in the cloud while exploiting low-latency drafting on the edge, the framework balances efficiency, accuracy, and cost in distributed inference.


\begin{algorithm}
\caption{Speculative Decoding for Distributed Inference}\label{alg:sd-distributed}
\begin{algorithmic}[1]
\State \textbf{Input:} Prompt $y_{1:m}$, Draft model $\mathcal{M}_q$ (edge), Verifier model $\mathcal{M}_p$ (cloud)
\State \textbf{Initialize:} $Y \gets [\,]$ \Comment{Generated sequence}

\While{$<eos>$ token not predicted}
    \State \textbf{Edge:} Generate $d$ speculative tokens
    \For{$i\in [1, d]$ }
    \State$y_{m+i}\gets\mathcal{M}_q(y_{1:m+i-1})\sim q(\cdot)$
    \EndFor
    \State \textbf{Send ${y_{m+1:m+d}}$ to cloud for verification}
    \State \textbf{Cloud:} In parallel, [$y_{m+i}^{*} \gets \mathcal{M}_p(y_{1:m+i-1}) \sim p(\cdot)$ for $i\in[1, d+1]$] in a single forward pass.
    \State Verify($y_{m+i}, y_{m+i}^{*}$) in parallel.
    \State Accept the longest verified draft ${y}_{m:m+k}$
    \State Sample $\tilde{y}_{m+k+1}\gets norm(max(0, p-q)$
    \State Correct the rejected draft $y_{m+k+1} = \tilde{y}_{m+k+1}$.
        \State Append tokens ${y}_{m:m+k+1}$ to $Y$
        \State Update context: $m\gets m+k+1$, $y_{1:m+k+1}$

\EndWhile

\State \textbf{Return:} $Y$
\end{algorithmic}
\end{algorithm}

\begin{table*}[ht]
\centering
\small
\caption{Mean performance and efficiency metrics on the \textbf{Spec-Bench} dataset, comparing \textit{Target}, \textit{Draft}, and \textit{Speculative} models across NLP tasks. Higher ($\uparrow$) is better performance, lower ($\downarrow$) indicates greater efficiency.}
\label{tab:specbench_results}
\begin{tabular}{lccc}
\toprule
\textbf{Task / Metric} & \textbf{Target Model} & \textbf{Draft Model} & \textbf{Our Model (\our{})} \\
\midrule
\multicolumn{4}{l}{\textbf{Task Scores} ($\uparrow$)} \\
\midrule
Multi-turn Conversation        & 3.62 & 2.67 & 3.52 \\
Translation                    & 4.80 & 3.85 & 4.71 \\
Summarization                  & 4.61 & 4.41 & 4.55 \\
Question Answering             & 4.25 & 3.08 & 4.15 \\
Mathematical Reasoning         & 4.88 & 3.19 & 4.68 \\
Retrieval-Augmented Generation & 4.56 & 3.13 & 4.68 \\
\midrule
\textbf{Overall Score} ($\uparrow$) & \textbf{4.45} & \textbf{3.39} & \textbf{4.38} \\
\midrule
\multicolumn{4}{l}{\textbf{Efficiency Metrics}} \\
\midrule
Mean Target Model Calls ($\downarrow$)     & 133.25 & 0.00   & 30.16 \\
Average Throughput (tokens/s) ($\uparrow$) & 2.43   & 3.91  & 3.25 \\
Cloud runtime ($\downarrow$) &  1.00$\times$  &  --  &  0.23$\times$ \\
\bottomrule
\end{tabular}
\end{table*}

\begin{table*}[ht]
\centering
\small
\caption{Performance comparison on the \textbf{CNN / DailyMail} summarization dataset. 
Higher values ($\uparrow$) indicate better generation quality.}
\label{tab:cnnDataset}
\begin{tabular}{lccc}
\toprule
\textbf{Metric} & \textbf{Target Model} & \textbf{Draft Model} & \textbf{Our Model (\our{})} \\
\midrule
BLEU-1 ($\uparrow$)      & 23.76 & 22.33 & 23.39 \\
BLEU-4 ($\uparrow$)      & 07.57  & 06.49  & 06.98  \\
ROUGE-1$_\text{F1}$ ($\uparrow$) & 38.38 & 36.05 & 37.99 \\
ROUGE-L$_\text{F1}$ ($\uparrow$) & 24.32 & 22.49 & 23.92 \\
CIDEr-D ($\uparrow$)     & 02.50  & 01.15  & 03.19  \\

\midrule
\multicolumn{4}{l}{\textbf{Efficiency Metrics}} \\
\midrule
Mean Target Model Calls ($\downarrow$)     & 127.30 & 0.00   & 30.79 \\
Average Throughput (tokens/s) ($\uparrow$) & 1.21   & 2.82  & 1.95 \\
Cloud runtime ($\downarrow$) &  1.00$\times$  &  --  &  0.24$\times$ \\
\bottomrule
\end{tabular}
\end{table*}

\textbf{Draft token length:}
The number of draft tokens, denoted as $d$, generated per edge call is a key control parameter in speculative decoding, governing the trade-off between computation and communication in the edge–cloud pipeline. A small $d$ increases verification frequency and synchronization overhead, while a large $d$ raises token rejection likelihood and draft computation cost, reducing overall speedup.

From a systems perspective, $d$ balances edge computation and cloud communication: smaller values make the system communication-bound, whereas larger values risk redundant local computation. Thus, $d$ is treated as a hyperparameter, selected empirically by monitoring acceptance rates on an initial validation subset (typically $\sim10$ samples). The objective is to maximize throughput while maintaining a stable acceptance rate under realistic latency and compute constraints.





\section{Experiments}

\textbf{Datasets:} We use publicly available datasets, including CNN/DailyMail \cite{see2017get} for summarization, and Spec-Bench \cite{xia2024unlocking}, a benchmark designed to evaluate speculative decoding methods. Spec-Bench consists of six subtasks—multi-turn conversation, summarization, translation, retrieval-augmented generation, question answering, and mathematical reasoning—allowing fair comparison of both speed and performance across diverse NLP settings.

\textbf{Setup:} Our framework follows an edge–cloud design, where a small draft model operates at the edge and a larger verification model runs on the cloud. The edge setup uses an NVIDIA RTX 3080 GPU (12 GB RAM), providing a balance between computational capability and deployability for high-end edge devices. The cloud setup employs an NVIDIA RTX A6000 GPU (48 GB RAM), representing typical industrial-scale infrastructure for large model inference.

\textbf{Models:} To instantiate this setup, we use \texttt{LLaMA-3.2-1B} as the draft model at the edge and \texttt{LLaMA-3.1-8B} \cite{dubey2024llama} as the target verification model on the cloud. This configuration reflects a realistic distributed scenario in which a lightweight model performs draft generation, while the larger model validates outputs to ensure correctness during verification.

\textbf{Metrics:} For the CNN/DailyMail dataset, we evaluate generated summaries using standard metrics. BLEU-1 and BLEU-4 assess n-gram precision, ROUGE-1 and ROUGE-L measure unigram recall and sequence-level overlap, and CIDEr-D evaluates semantic relevance using TF-IDF–weighted n-grams. All metric scores are normalized to a 0–100 scale for consistency.

For the Spec-Bench dataset, performance is evaluated using \texttt{Gemini-2.5-Flash-Lite} \cite{comanici2025gemini} as an automatic judge, which assigns each generated response a score on a 1-5 Likert scale. Responses are assessed across six dimensions, correctness, instruction-following, completeness, clarity and coherence, conciseness, and factuality, with higher scores indicating better overall performance.

\textbf{Baseline:} We compare against two baselines: (1) the \textit{Target Model}, a large cloud model representing the upper bound in output quality, and (2) the \textit{Draft Model}, a lightweight edge model providing low-latency outputs. (3) Our method \textit{\our{}} bridges these extremes using speculative decoding, where the target model selectively verifies draft outputs.

\section{Results and Analysis}

\begin{figure}
    \centering
    \includegraphics[scale = 0.25]{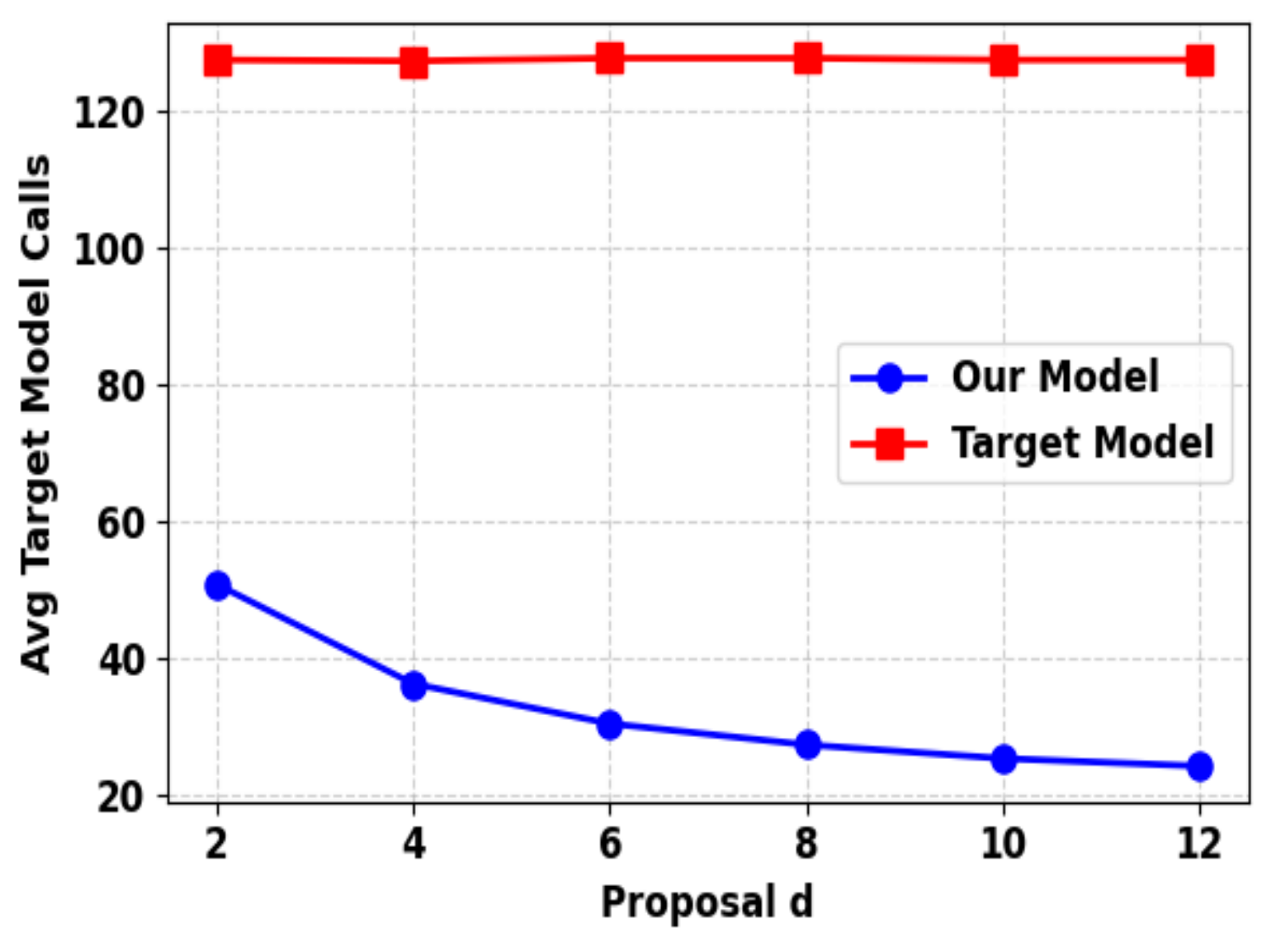}
    \caption{Performance trade-off of our method, illustrating the variation in target model calls when the hyperparameter $d$ (speculative token generation before one verifier call) increases.}
    \label{fig:d_vs_tgtcalls}   
\end{figure}

In Table \ref{tab:specbench_results}, we present results of our method on the SpecBench dataset, alongside two baselines: the Full Model (fully deployed on the cloud) and the Draft Model (operating solely at the edge). Evaluation uses an LLM-as-a-judge framework across multiple qualitative dimensions, correctness, instruction-following, completeness, clarity and coherence, conciseness, and factuality, with average scores reported as the overall performance metric.

In addition to performance, we evaluate system efficiency using three indicators: average throughput (tokens per second), average number of cloud model calls, and average cloud runtime reduction. Our method achieves performance comparable to the full model while reducing cloud model calls by 77.4\%, yielding significant efficiency gains with minimal quality degradation.

Table \ref{tab:cnnDataset} reports results on the CNN/DailyMail dataset using standard metrics (BLEU-1, BLEU-4, ROUGE-1, ROUGE-L, CIDEr-D) to measure linguistic quality and relevance, along with the same efficiency metrics. The results show that our method closely matches the accuracy and fluency of the full model, while reducing cloud calls by 76\% and substantially lowering computational cost.

Overall, these findings demonstrate that our approach maintains near cloud-level performance while significantly improving efficiency. Consistent results across SpecBench and CNN/DailyMail highlight its robustness and generalizability, supporting the use of selective, hybrid edge–cloud inference.

\textbf{Analysis on the value of $d$:}
In Figure~\ref{fig:d_vs_tgtcalls}, we analyze the effect of the number of draft tokens ($d$) on target model calls for CNN/DailyMail. Increasing $d$ consistently reduces target model invocations, as more draft tokens decrease verification frequency. This effect is amplified by strong alignment between the draft and target models, leading to higher token acceptance rates. Poor alignment, however, would reduce acceptance and increase target calls.

\section{Conclusion}
We propose a distributed inference framework, \our{}, that leverages speculative decoding in a dual-model setup, with a lightweight draft model at the edge and a larger verification model on the cloud. By using the verifier only for parallel verification rather than autoregressive generation, \our{} reduces costly cloud invocations. Experiments across multiple NLP tasks show that \our{} significantly lowers cloud model calls while maintaining performance close to the full model. These results establish \our{} as a practical, scalable solution for latency-aware distributed inference, effectively balancing efficiency without any loss in performance.

\bibliographystyle{IEEEtran}
\bibliography{mab}

\end{document}